\documentclass[10pt,twocolumn,letterpaper]{article}

\usepackage[pagenumbers]{cvpr} 

\usepackage[accsupp]{axessibility}  
\usepackage{multicol}
\usepackage{multirow}
\usepackage{makecell}
\usepackage{booktabs}
\usepackage{tabu}
\usepackage{pifont}
\usepackage{ifthen}
\usepackage{graphicx}
\usepackage{caption}
\definecolor{cvprblue}{rgb}{0.21,0.49,0.74}
\usepackage[pagebackref,breaklinks,colorlinks,allcolors=cvprblue]{hyperref}

\def\paperID{11610} 
\def\confName{CVPR}
\def\confYear{2025}

\title{\textit{Cholec}{Track20}: A Multi-Perspective Tracking Dataset for Surgical Tools}

\author{Chinedu Innocent Nwoye$^{1,3}$ \quad  Kareem Elgohary$^{1}$ \quad Anvita Srinivas$^{1}$\\Fauzan Zaid$^{1}$ \quad Joël L. Lavanchy$^{2}$ \quad  Nicolas Padoy$^{1,3}$\vspace{0.3em} \\
{\normalsize $^1$University of Strasbourg, CNRS, INSERM, ICube, UMR7357, Strasbourg, France} \\
{\normalsize $^2$University of Basel, University Digestive Health Care Center, Clarunis, Switzerland}\\
{\normalsize $^3$IHU Strasbourg, France} \\
{\small{Project page:~\tt \url{https://github.com/camma-public/cholectrack20}}}
}

\begin{document}

\twocolumn[{%

\vspace*{-16mm}
\begin{center}
\fboxsep=3pt
\colorbox[cmyk]{1.0,0.8,0.0,0.0}{\parbox{0.95\linewidth}{%
    \centering\color{white}\small\bfseries%
    Peer-Reviewed and Accepted at CVPR 2025%
}}
\end{center}
\vspace{4mm}

\renewcommand\twocolumn[1][]{#1}%
\maketitle

\begin{center}
    \centering
    \captionsetup{type=figure}
    \includegraphics[width=\textwidth,height=8.002cm]{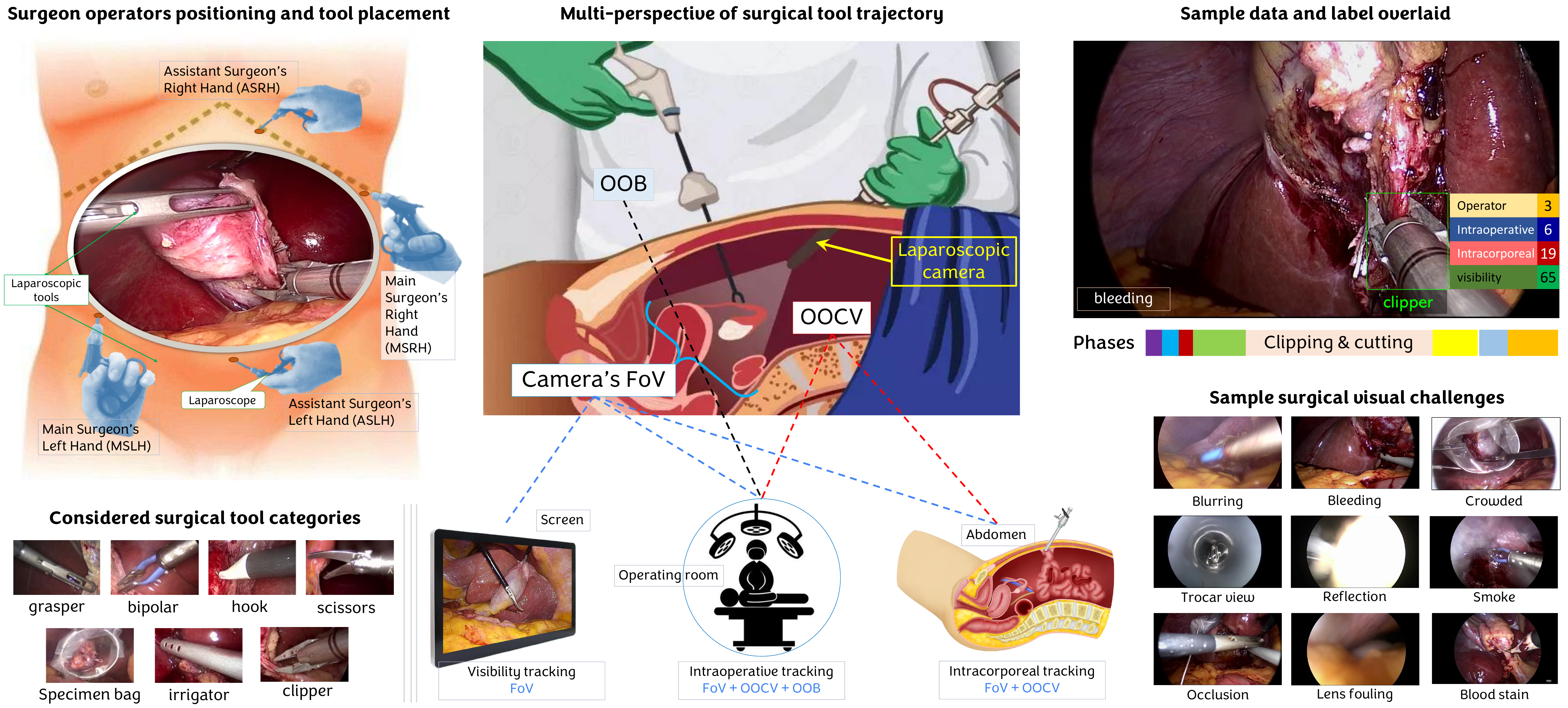}
    \captionof{figure}{Illustration of multi-perspective tracking in surgical domain and CholecTrack20 dataset labels for surgical tool tracking.}
    \label{fig:multi_perspective}
\end{center}%
}]

\begin{abstract}
Tool tracking in surgical videos is essential for advancing computer-assisted interventions, such as skill assessment, safety zone estimation, and human-machine collaboration. However, the lack of context-rich datasets limits AI applications in this field. Existing datasets rely on overly generic tracking formalizations that fail to capture surgical-specific dynamics, such as tools moving out of the camera’s view or exiting the body. This results in less clinically relevant trajectories and a lack of flexibility for real-world surgical applications. Methods trained on these datasets often struggle with visual challenges such as smoke, reflection, and bleeding, further exposing the limitations of current approaches.
We introduce CholecTrack20, a specialized dataset for multi-class, multi-tool tracking in surgical procedures. It redefines tracking formalization with three perspectives: (1) intraoperative, (2) intracorporeal, and (3) visibility, enabling adaptable and clinically meaningful tool trajectories. The dataset comprises 20 full-length surgical videos, annotated at 1 fps, yielding over 35K frames and 65K labeled tool instances. Annotations include spatial location, category, identity, operator, phase, and scene visual challenge.
Benchmarking state-of-the-art methods on CholecTrack20 reveals significant performance gaps, with current approaches ($<45\%$ HOTA) failing to meet the accuracy required for clinical translation. These findings motivate the need for advanced and intuitive tracking algorithms and establish CholecTrack20 as a foundation for developing robust AI-driven surgical assistance systems.
The dataset is released under CC-BY-NC-SA 4.0 license and is available for download through the project page.\\
\vskip-3.8em
\end{abstract}
\section{Introduction}
\label{sec:intro}
The true impact of computer vision research lies in its practical applications, especially in critical fields like healthcare. Among these, surgery represents one of the most demanding domains, providing a definitive test for the capabilities of vision technologies.
Surgical data science, on its own, has significantly advanced interventional healthcare by leveraging data-driven techniques to provide critical decision support to medical professionals \cite{maier2017surgical}. A key area of this advancement is the analysis of endoscopic video data, which offers real-time insights into surgical procedures, aids in skill assessment, and helps predict complications \cite{twinanda2016endonet,czempiel2020tecno,xu2022information}. Accurate tracking of surgical tools is central to these analyses, as it guides temporal progression of procedural phases and correlates with surgical actions and management of adverse events~\cite{nwoye2020recognition,wagner2023comparative}.
Despite progress in computer vision, research~\cite{ramesh2023dissecting} shows that deep learning models pretrained on general datasets often struggle with surgical contexts due to complex scene dynamics, diverse tool types, and challenging visual conditions such as bleeding, smoke, and variable lighting \cite{geiger2012we, dendorfer2020mot20, sun2022dancetrack}.
This highlights the need for domain-specific datasets tailored to the unique requirements of surgical tool tracking.
Obtaining medical and surgical data for research is challenging due to ethical and practical constraints, and annotating it requires expert knowledge.
Current methods for tool tracking largely focus on Single Object Tracking (SOT) \cite{zhao2019surgical}, Multi-Class Tracking (MCT) with one tool per class \cite{nwoye2019weakly,nwoye2021deep}, or Multi-Object Tracking (MOT) treating all tools as a single class \cite{robu2021towards,fathollahi2022video}. However, these approaches often miss the complexities of Multi-Class Multi-Object Tracking (MCMOT) specific to surgical contexts, where tools interact dynamically and may move out of the camera's field of view or within the body cavity.

\textbf{Multi-perspective tracking} addresses these challenges by defining tool trajectories across different viewpoints during surgical procedures. It includes three critical perspectives (illustrated in \cref{fig:multi_perspective}): (1) \textit{intraoperative} - covering the entire procedure duration to monitor tool usage and assess surgical proficiency; (2) \textit{intracorporeal} - focusing on tool tracks within the body to evaluate specific tasks and predict risks; and (3) \textit{visibility} - tracking tools within the camera's field of view to provide real-time feedback to surgeons.
Existing surgical tracking datasets~\cite{sznitman2012data,gao2014jhu} often lack these levels of granularity and adaptability needed for comprehensive tool modeling. They generally follow generic tracking formalizations and struggle to capture the intricacies of surgical tool interactions, particularly when tools are replaced or move beyond camera visibility \cite{sznitman2012data, bouget2017vision,lee2019weakly}.

To address this gap, we introduce \textit{CholecTrack20}, a novel dataset, based on laparoscopic cholecystectomy surgery, designed for multi-class multi-tool tracking from intraoperative, intracorporeal, and visibility perspectives. Derived from raw laparoscopic videos \cite{twinanda2016endonet,nwoye2022rendezvous}, it includes detailed annotations such as spatial coordinates, tool categories, track identities (IDs), visual challenges, phase labels, and other scene attributes. This dataset enhances benchmarking resources for computer vision research and supports the development of AI models tailored to surgical tool tracking, phase recognition, and surgeon performance assessment.
This paper provides a comprehensive overview of the data acquisition and annotation methodology, detailed data analysis, and technical validation. 
In addition, we conduct extensive benchmark experiments using state-of-the-art deep learning methods for object detection and tracking, evaluating performance across various surgical phases and visual challenges, discussing insightful findings. 

In summary, the main contributions are:

\begin{enumerate}
    \item Introduction of \textit{CholecTrack20}, a pioneering dataset for multi-perspective tracking with extensive annotations.
    \item Extensive experimental analysis validating the dataset's effectiveness for surgical tool detection and tracking.
    \item Insights from model performance analysis under diverse visual challenges, highlighting the utility of each tracking perspective for AI-driven surgical solutions.
\end{enumerate}

\section{Related Works}
\label{sec:literature}
\noindent\textbf{Object detection and tracking.~}
Advances in object detection and tracking have been driven by datasets like COCO \cite{lin2014microsoft}, KITTI \cite{geiger2012we}, MOTChallenge \cite{dendorfer2020mot20}, VisDrone \cite{zhu2021detection}, DanceTrack \cite{sun2022dancetrack}, and TAO \cite{dave2020tao}, enabling progress in Single Object Tracking (SOT) \cite{danelljan2017eco}, Multi-Object Tracking (MOT) \cite{wojke2017deepsort,zhang2021fairmot}, and Multi-Class Multi-Object Tracking (MCMOT) \cite{lee2016multi,du2021giaotracker}.
Applying these techniques to surgical tool tracking poses challenges especially in the phase of bleeding, smoke, rapid movements, and variable lighting. Traditional tracking, centered on visibility, struggles when tools leave the camera’s view or are replaced during surgery.\smallskip

\noindent\textbf{Surgical tool tracking and configurations.~}
While electromagnetic and optical tracking methods \cite{fried1997image,chmarra2007systems} have been explored in surgical domain, image-based approaches \cite{sznitman2012data} better align with surgeons’ view but face issues like identity fragmentation, identity switch, and low tracking accuracy \cite{bouget2017vision,lee2019weakly, nwoye2019weakly,robu2021towards, nwoye2021deep,fathollahi2022video} especially with mid-procedure tool replacements and tools exiting/re-entering the field of view or body cavity. 
To fully capture tool usage complexity, it is crucial to consider multiple perspectives on tool trajectories. Existing datasets \cite{qiu2019real, fathollahi2022video} provide insights but lack comprehensive coverage of these scenarios, highlighting the need for detailed datasets addressing these complexities.
\section{Methodology}
\label{sec:methods}
CholecTrack20 is a detailed dataset for surgical tool tracking in laparoscopic cholecystectomy, offering binary and spatial annotations for tools, including identity, category, bounding box location, motion, operator, phase, activity, usage conditions, and visual challenges, essential for training and benchmarking surgical AI tool tracking algorithms.

\subsection{Data Acquisition and Collection}
{\bf Data source.}
The raw videos are sourced from the publicly available Cholec80 \cite{twinanda2016endonet} and CholecT50 \cite{nwoye2022rendezvous} datasets, with appropriate permissions and adherence to license terms. Recorded at the University Hospital of Strasbourg  with the aid of laparoscopic cameras, these videos document laparoscopic cholecystectomy surgeries.\smallskip

\noindent{\bf Video selection.}
Long videos were systematically chosen to represent surgical complexities and tool variability, capturing all key phases of laparoscopic cholecystectomy. High-quality videos with clear visuals were selected, subsampled from 25 FPS to 1 FPS for annotation, maintaining temporal consistency and clarity.\smallskip

\noindent{\bf Sensitive data handling.}
To protect patient and staff identities, out-of-body frames potentially revealing sensitive information, such as the identity of patients, clinical staff, or the operating room, were checked and anonymized following established techniques \cite{lavanchy2023preserving}. This ensures compliance with ethical standards and privacy regulations.


\subsection{Track Formalization}
Given a video dataset $D = \{S_1, S_2, \ldots, S_n\}$ of laparoscopic surgeries, each sequence $S_i$ includes frames annotated with bounding boxes $B = [B_1, B_2, \ldots, B_M]$ for tool locations and classes $C = \{C_1, C_2, \ldots, C_N\}$. Tools, manipulated by operators linked to trocar ports $P = [P_1, P_2, \ldots, P_M]$, are assigned unique track identities (ID) through time. The $\text{ID}$ reassignment is guided by visual cues and clinical knowledge. Visually, it is based on class $c \in C$ and location $b \in B$. Contextually, it considers the role and hand position of the surgeon operator linked to a unique trocar port $p \in P$.
Tool tracking solves the association matrix $A(t)$, where $A_{i,j}=1$ indicates the $i$-th tool in frame $t$ is associated with the $j$-th tool in frame ${t+1}$, and $A_{i,j}=0$ otherwise. The aim is to obtain tool trajectories $T = \{T_1, T_2, ..., T_K\}$, each uniquely identified by $\text{ID}$, taking into consideration the visibility, intracorporeal, and intraoperative use.

\subsection{Multi-Perspective (MP) Trajectory}
\label{sec:trajectory}
Defining tool trajectories in surgical procedures necessitates a unique approach, given the variability in tracking across different perspectives, formalized as follows:

\begin{enumerate}[wide, labelwidth=!, labelindent=0pt]
    \item \textbf{Intraoperative trajectory}: This lifelong tracking starts with a tool's first appearance and ends at its last in a patient's body during a procedure. It requires re-identification post-occlusion, out-of-camera view, or reinsertion. This approach is vital for applications such as tool usage monitoring \cite{al2018monitoring}, inventory management \cite{moatari2017improving}, surgeon training \cite{lee2020evaluation}, skill assessment \cite{fathollahi2022video}, and tool usage pattern analysis \cite{sun2010estimation}.


    \item \textbf{Intracorporeal trajectory}: Unique tracks begin when a tool enters the body and end when it exits through a trocar port, even if off-camera. If a tool exits outside the camera’s view, the exit is inferred if another tool enters through the same trocar or the initial tool releases its grasp. This is essential for understanding surgical workflow, as some actions occur outside the camera's focus, like graspers holding tissue out of view to facilitate other tools' actions \cite{lee2020evaluation}. Intracorporeal tracking supports a range of AI tasks including action evaluation \cite{toor2022optimizing}, skill assessment \cite{fathollahi2022video, lee2020evaluation}, tool usage optimization \cite{richa2011visual}, and surgical risk estimation \cite{sun2010estimation}.

    \item \textbf{Visibility trajectory}: 
    Tracking starts with a tool's first appearance within the camera view and ends when it leaves the view.
    It requires re-identification (re-ID) after occlusions
    or brief periods of disappearance within a two-second tolerance.
    This method is useful for assessing surgeon actions \cite{lee2020evaluation} and skill training \cite{dubin2018model,lavanchy2021automation}, providing go-no-go decision support, and measuring economy of motion \cite{shbool2023economy}.
\end{enumerate}

\begin{figure}[t]
    \centering
    \includegraphics[width=.95\columnwidth]{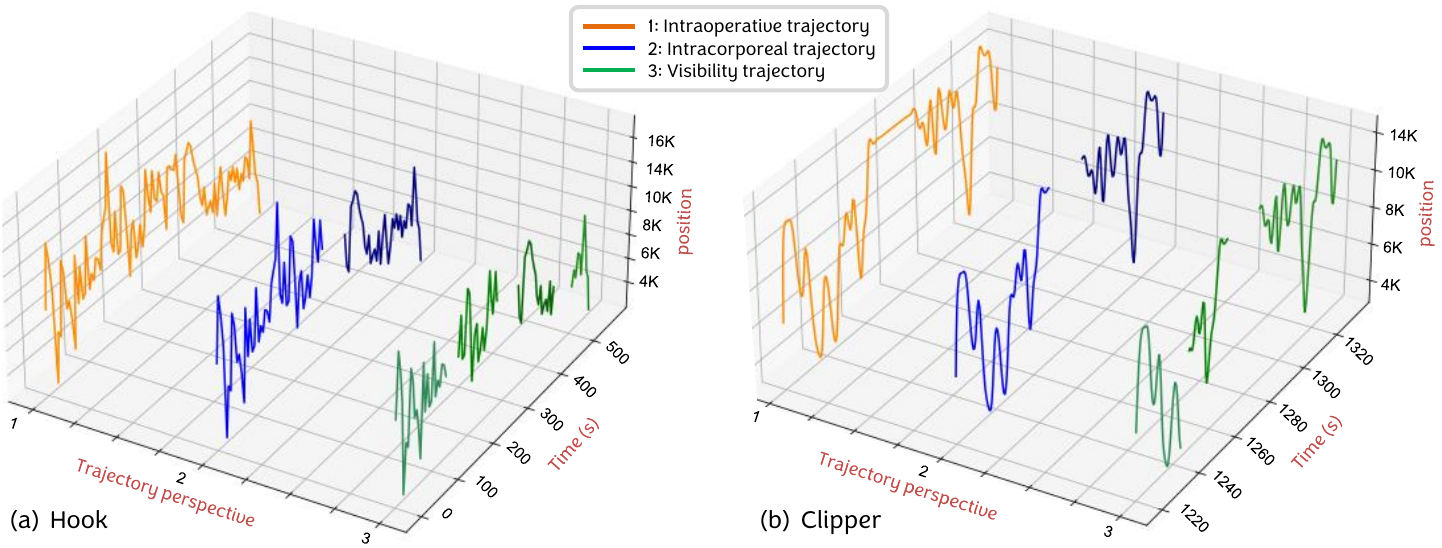}
    \caption{
    Multi-perspective trajectories of surgical tool.
    }
    \label{fig:multi-p}
\end{figure}

Some existing studies \cite{nwoye2019weakly,fathollahi2022video} follow the intraoperative tracking format, others \cite{robu2021towards,lavanchy2021automation} employs visibility trajectory. The intracorporeal trajectory, being the most complex to annotate, are not well-represented in the literature. Our dataset is the first to provide fine-grained labels for all the three (\cref{fig:multi-p}), enabling statistical analyses such as tool usage counts, events counts, abdominal entries/exits, tool idleness, out-of-camera view occurrences, and mean tracklets per perspective.

\subsection{Data Annotation}
\label{sec:data-annotation}

\noindent{\bf Annotators and tools.}
Bounding boxes were annotated by four researchers skilled in surgical workflow analysis, supplemented by prior study data \cite{nwoye2019weakly,vardazaryan2018weakly,nwoye2023cholectriplet2022}. Annotation tools used include {\it Annonymized System},
and a custom Python tool for visualization, merging, and validation. A pre-designed guide, refined by surgical experts, described the labels and provided image guidance as needed.\smallskip

\noindent{\bf Label types and categories.}
Tool spatial positions were annotated using bounding box coordinates, while other labels were represented by class identities.
Seven predominant tool categories were defined: (1) \textit{cold} grasper, (2) bipolar \textit{grasper}, (3) \textit{monopolar} hook, (4) \textit{monopolar} scissors, (5) clipper or \textit{clip applier}, (6) irrigator or \textit{suction device}, and (7) \textit{specimen} bag.
Four operator categories were defined: (1) main surgeon left hand \textit{MSLH}, (2) main surgeon right hand \textit{MSRH}, (3) assistant surgeon right hand \textit{ASRH}, and (4) null operator \textit{NULL}. The assistant surgeon left hand \textit{ASLH} holding the endoscopic camera is unreported.
A total of eight visual challenges were noted: (1) blurring, (2) bleeding, (3) camera lens fouling, (4) crowded scene, (5) occlusion, (6) smoke, (7) specular light reflection, and (8) trocar view or under-coverage.
The seven commonest surgical phases were annotated: (1) preparation, (2) calot triangle dissection, (3) gallbladder dissection, (4) clipping \& cutting, (5) gallbladder packaging, (6) cleaning and coagulation, and (7) gallbladder extraction. \smallskip

\begin{figure}[t]
    \centering
    \includegraphics[width=.95\columnwidth]{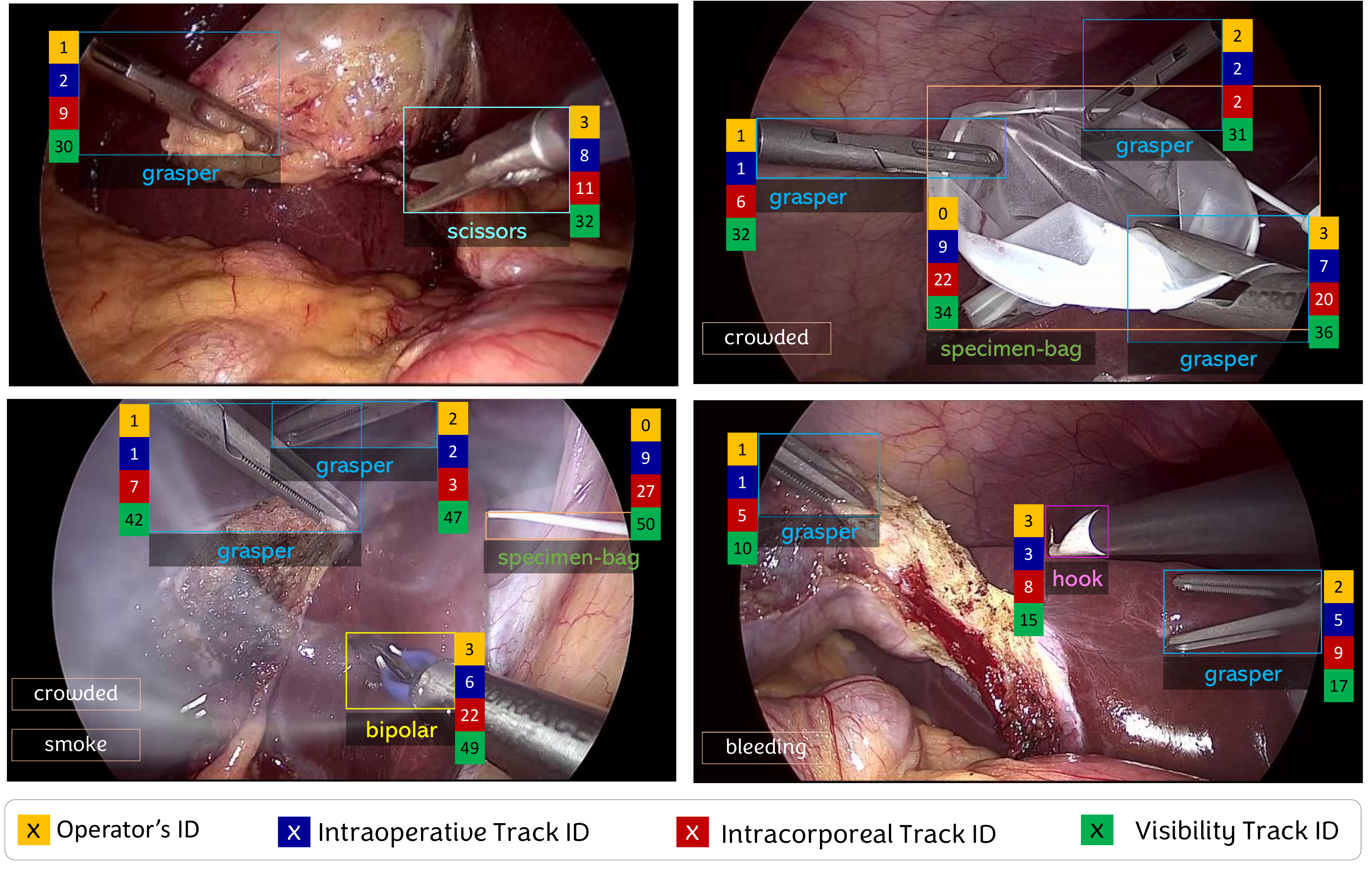}
    \caption{Examples of images from CholecTrack20 tracking dataset with the labels overlaid on the raw images.}
    \label{fig:data-samples}
\end{figure}

\noindent{\bf Annotation process.}
Annotations involved drawing bounding boxes $[x,y,w,h]$ over tooltips tagged with tool class $c \in C$ and operator class $p \in P$, following trocar ports for accurate surgeon identification. Surgical details such as phase, out-of- view statuses (camera/abdomen), tool entry/exit, and visual challenge attributes were annotated to aid accurate track assignment. Tool-tissue interaction labels from CholecT50~\cite{nwoye2022rendezvous} provides additional help in perpetuating track identities. Annotations were reviewed at 25 FPS in uncertain cases, and underwent rigorous quality control, ensuring high-quality labels over two years.
\cref{fig:data-samples} presents samples of images from the CholecTrack20 dataset alongside their respective annotations illustrating the meticulous labeling system employed to ensures a rich dataset for detailed surgical tool tracking analysis.

\subsection{Quality Assurance}
\noindent{\bf Label agreement.}
Two label agreement metrics validate dataset quality: Jaccard Index for spatial overlap of bounding boxes and Cohen’s Kappa Statistic for category labels.
The findings of our three validation forms include:

\begin{enumerate}
    \item \textit{Intra-rater agreement} involves self-correction. We observe a Jaccard Index of $99.4\%$ and Cohen’s Kappa Scores of $94.6\%$ for tools and $94.0\%$ for operators. 
    \item \textit{Inter-rater agreement} is evaluated on $20$ random samples across raters. The Jaccard Index is $91.8\%$, while tool class labels achieve $95.2\%$ and operator labels $92.7\%$ Cohen’s Kappa. Minor differences reflect high-quality annotations.
    \item \textit{Label mediation} uses a board-certified surgeon for ambiguities, particularly in operator labels. Out of $758$ uncertain samples, $133$ needed correction post-mediation.
\end{enumerate}

\begin{table*}[t]
    \centering
    \caption{\textbf{Dataset comparison} showing the scope, statistics, and attributes.
    Dataset marked $\ddag$ are not full-length videos.
    }
    \label{tab:comparison}
    \setlength{\tabcolsep}{3pt}
    \resizebox{0.95\linewidth}{!}{%
    \begin{tabular}{lccccrcccccrccc}
        \toprule
        \multirow{2}{*}{Dataset}&
        \multirow{2}{*}{Task}&
        \multicolumn{3}{c}{Track Perspectives}&\phantom{abc}&
        \multicolumn{5}{c}{Statistics}&\phantom{abc}&
        \multicolumn{3}{c}{Attributes} \\
        \cmidrule{3-5} \cmidrule{7-11} \cmidrule{13-15}
        && Visibility & Intraoperative & Intracorporeal &
        & \makecell{No. of\\Videos} & \makecell{Total\\Duration (s)} & \makecell{Frame\\Count} &
        \makecell{Tool\\Boxes} & \makecell{No. of\\Trajectories} & 
        & \makecell{Surgoen\\Operator} & \makecell{Visual\\Challenge} & \makecell{Surgical\\Phase}\\
        \midrule\midrule
        
        ATLAS Dione \cite{sarikaya2017detection}~$^\ddag$ & Detection &\ding{52}&\ding{56}&\ding{56} && 99 & - & 22467 & - & - && \ding{56}&\ding{56}&\ding{56} \\ 

        Cholec80-locations \cite{shi2020real} & Detection &\ding{52}&\ding{56}&\ding{56} && - & 4011 & 4011 & 6471 & - && \ding{56}&\ding{56}&\ding{56} \\

        Bouget et.al \cite{bouget2015detecting} & Detection &\ding{52}&\ding{56}&\ding{56} && 14 & - & 2476 & 3819 & - && \ding{56}&\ding{56}&\ding{56} \\

        m2cai16-tool-locations \cite{jin2018tool} & Detection & \ding{52}& \ding{56}& \ding{56} & & - & - & 2532 & 3038 & - && \ding{56}& \ding{56}& \ding{56} \\
        
        
        EndoVis`15 \cite{EndoVisSubInstrument}~$^\ddag$ & Tracking & \ding{52} & \ding{56} & \ding{56} && 16 & 540 & 13500 & - & - && \ding{56}&\ding{56}&\ding{56} \\

        Fathollahi et el \cite{fathollahi2022video}~$^\ddag$ & Tracking &\ding{56}& \ding{52}& \ding{56}&& 15 & 2700 & 2700 & - & - && \ding{56} & \ding{56} & \ding{56} \\

        RMIT \cite{sznitman2012data} & Detection \& Tracking & \ding{52} &\ding{56}&\ding{56}& & 4 & & 1500 & 1171 &  && \ding{56} & \ding{56} & \ding{56} \\
        
        CholecTrack20 (Ours) & Detection \& Tracking &\ding{52}& \ding{52}& \ding{52}&& 20 & 50581 & 35000 & 65200 & 2,624 && \ding{52}& \ding{52}& \ding{52}\\
        
        \bottomrule
    \end{tabular}
    }    
\end{table*}

\begin{figure}[!t]
    \centering
    \includegraphics[width=.98\columnwidth]{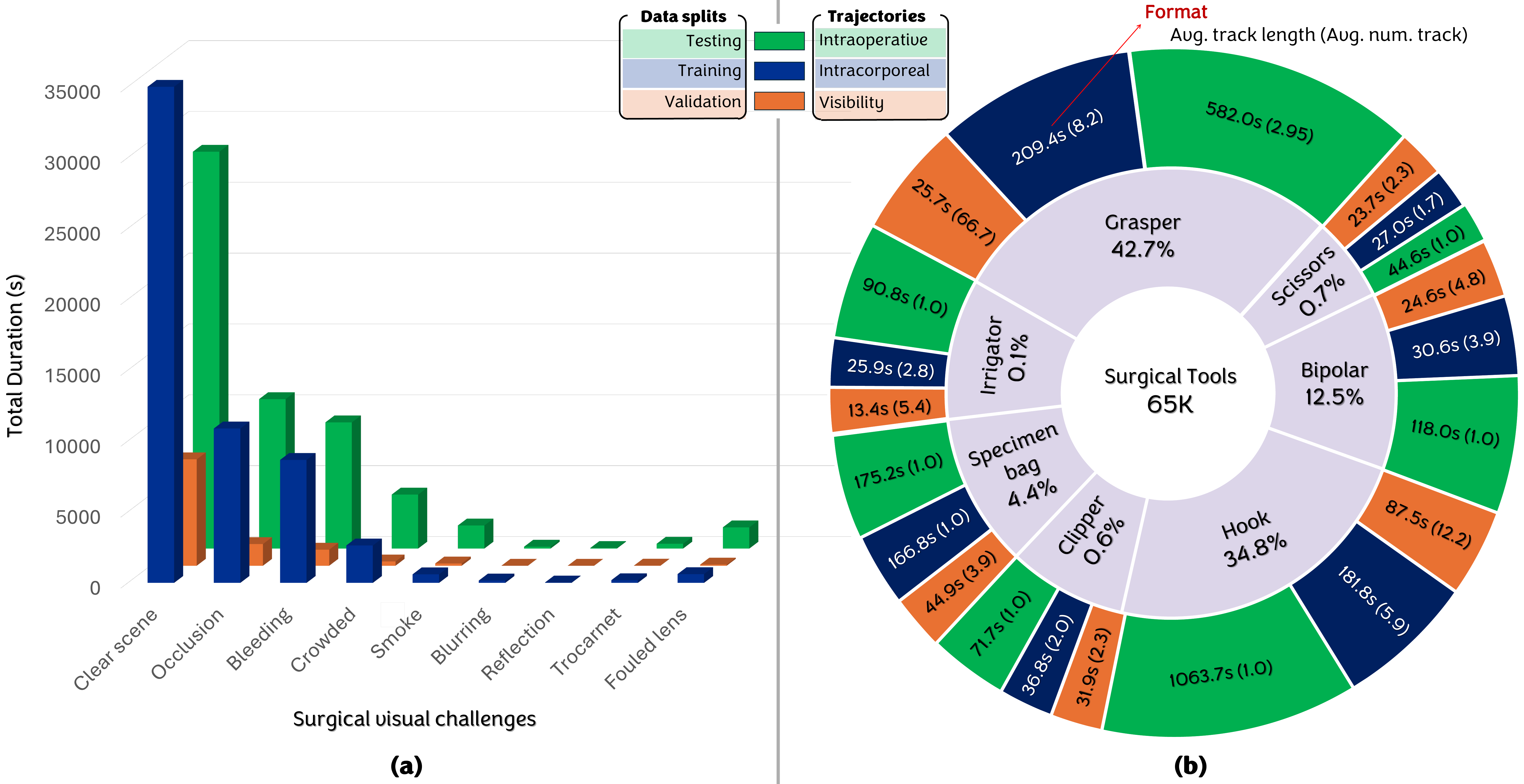}
    \caption{\textbf{Dataset statistics} on the distributions of (a) surgical scene visual challenges across data splits (b) track labels across perspectives, averaged across videos. Track length in seconds.}
    \label{fig:label-stat}
\end{figure}

\subsection{Data Statistics}


\noindent{\bf Quantitative overview.} The dataset includes 20 surgical videos totaling over 14 hours, averaging 42 minutes per surgery. As detailed in \cref{tab:comparison}, annotations cover 35,000 frames at 1 FPS ($\sim875,000$ at 25 FPS), with 65,000 bounding box labels, averaging two tools per frame. \cref{fig:label-stat}(a) summarizes track configurations across perspectives, showing tool usage by type and trajectory. For example, graspers average 27 minutes 39 seconds inside the body and are re-inserted 8.4 times, while irrigators are used for about 1 minute 30 seconds per surgery. The visibility perspective contributes approximately 2,000 trajectories. \cref{fig:label-stat}(b) outlines tool attributes and visual challenges in the dataset, with occlusion as most prevalent occurring up to 23,000 seconds, bleeding totals around 18,700 seconds. 
This analysis provides insights into the varied visual challenges encountered in laparoscopic cholecystectomy. Evaluating tracking methods across these challenges will reveal their strengths and weaknesses. Statistical details across dataset splits are presented in \cref{fig:label-stat}(b).\smallskip

\noindent{\bf Dataset comparison.} Existing publicly available datasets primarily focus on single perspective trajectory \cite{robu2021towards,fathollahi2022video}. In contrast, our dataset introduces a novel approach by annotation different perspectives, as presented in \cref{tab:comparison}, including visual challenges, phase details, activity labels, etc.
\smallskip

\noindent{\bf Dataset split.}
The dataset is split at the procedure level into non-overlapping training, validation, and testing sets in a 5:1:4 ratio, preventing data leakage. Video distribution is balanced by procedure duration to ensure similar complexity and difficulty levels across all splits.

%

\subsection{Data Analysis}
\begin{figure}[!tbp]
    \centering
    \includegraphics[width=.95\columnwidth]{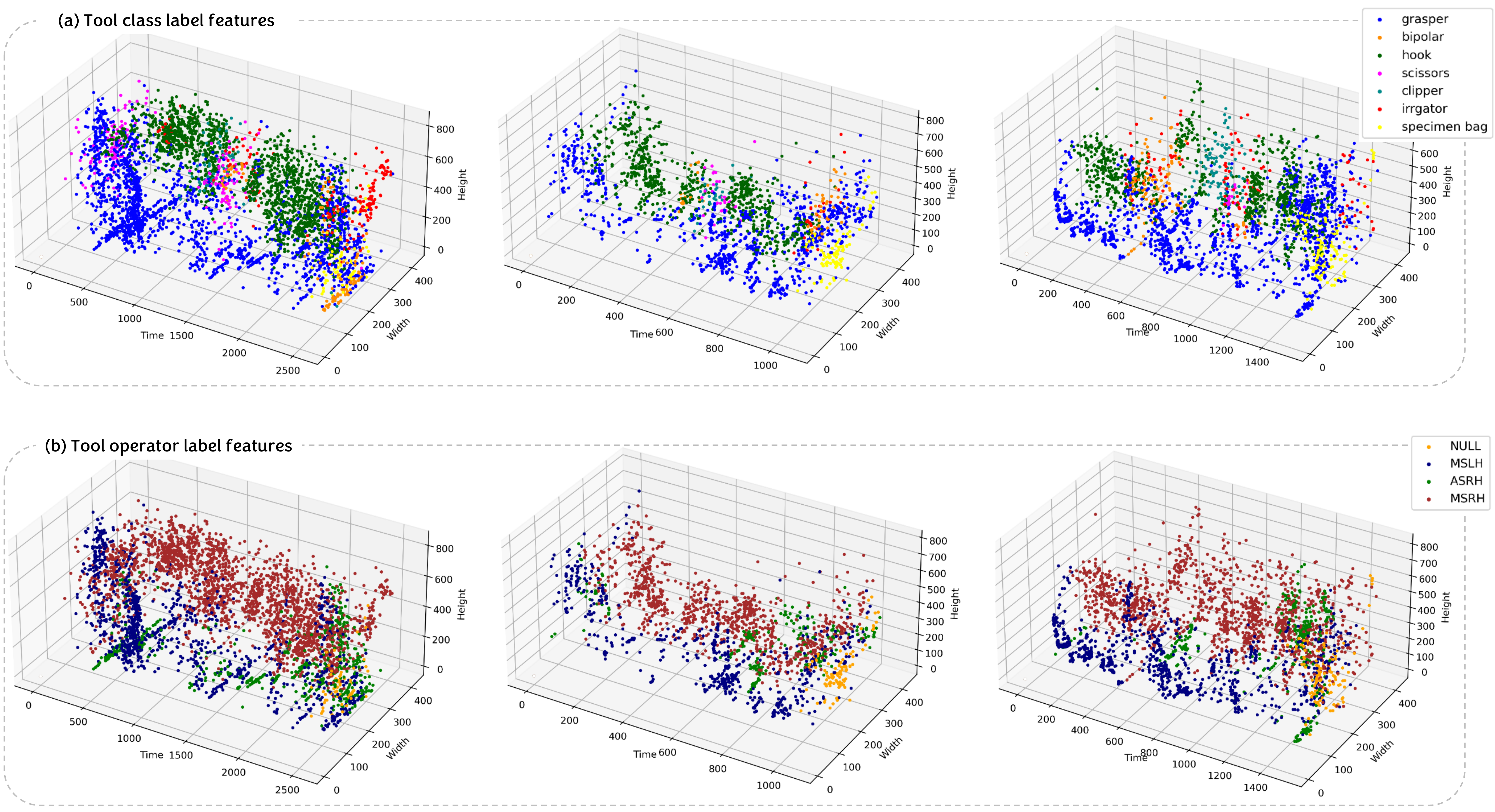}
    \caption{3D visualization of label alignments showing the tool position over track time. The coloring is for grouping features according to: (a) tool classes and (b) tool operators.}
    \label{fig:label-alignment}
\end{figure}

\begin{figure}[!tbp]
    \centering
    \includegraphics[width=.95\columnwidth]{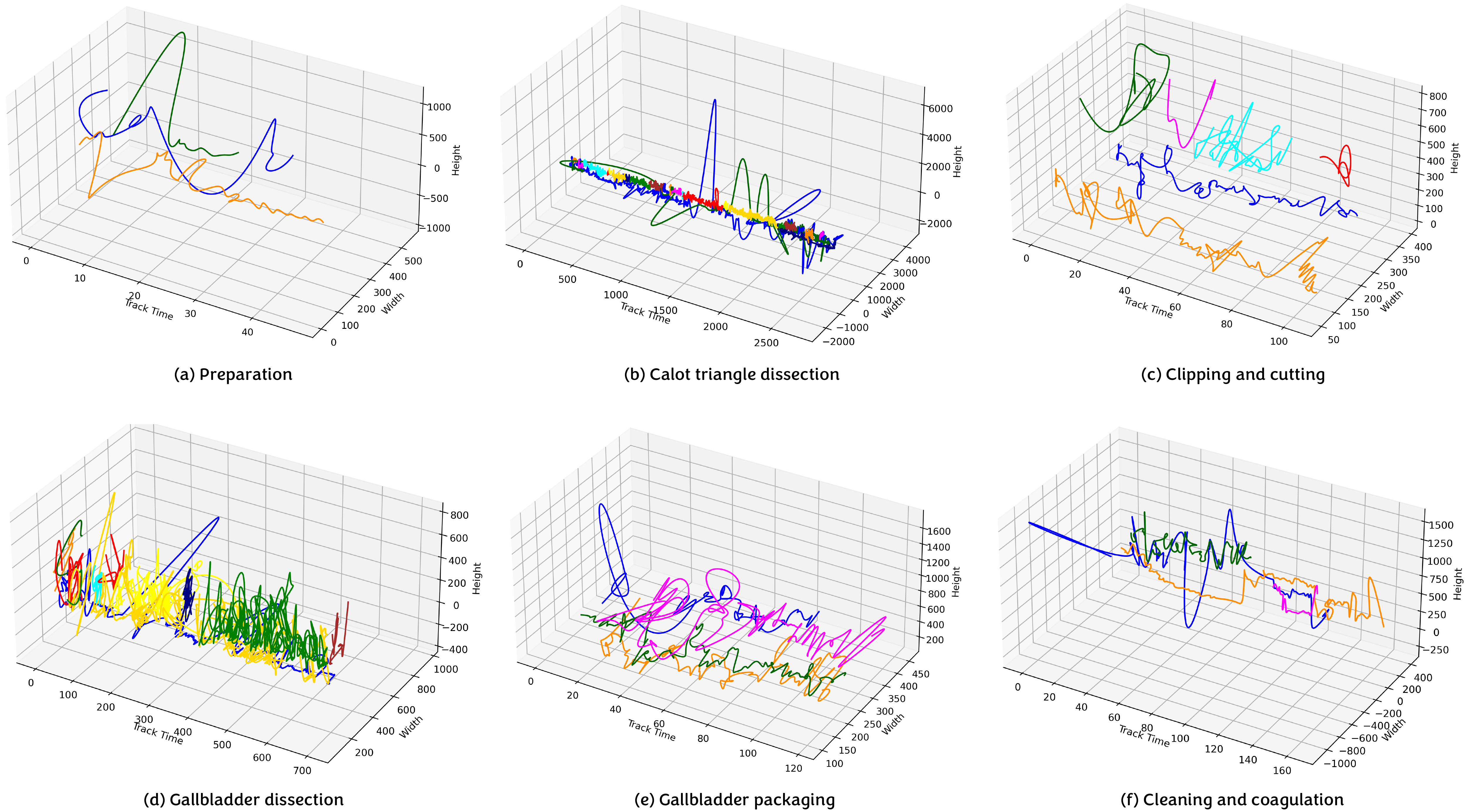}
    \caption{3D visualization of tracking across different surgical phases for some randomly selected videos.}
    \label{fig:phase-tracking}
\end{figure}

\begin{figure}[!tp]
    \centering
    \includegraphics[width=.95\columnwidth]{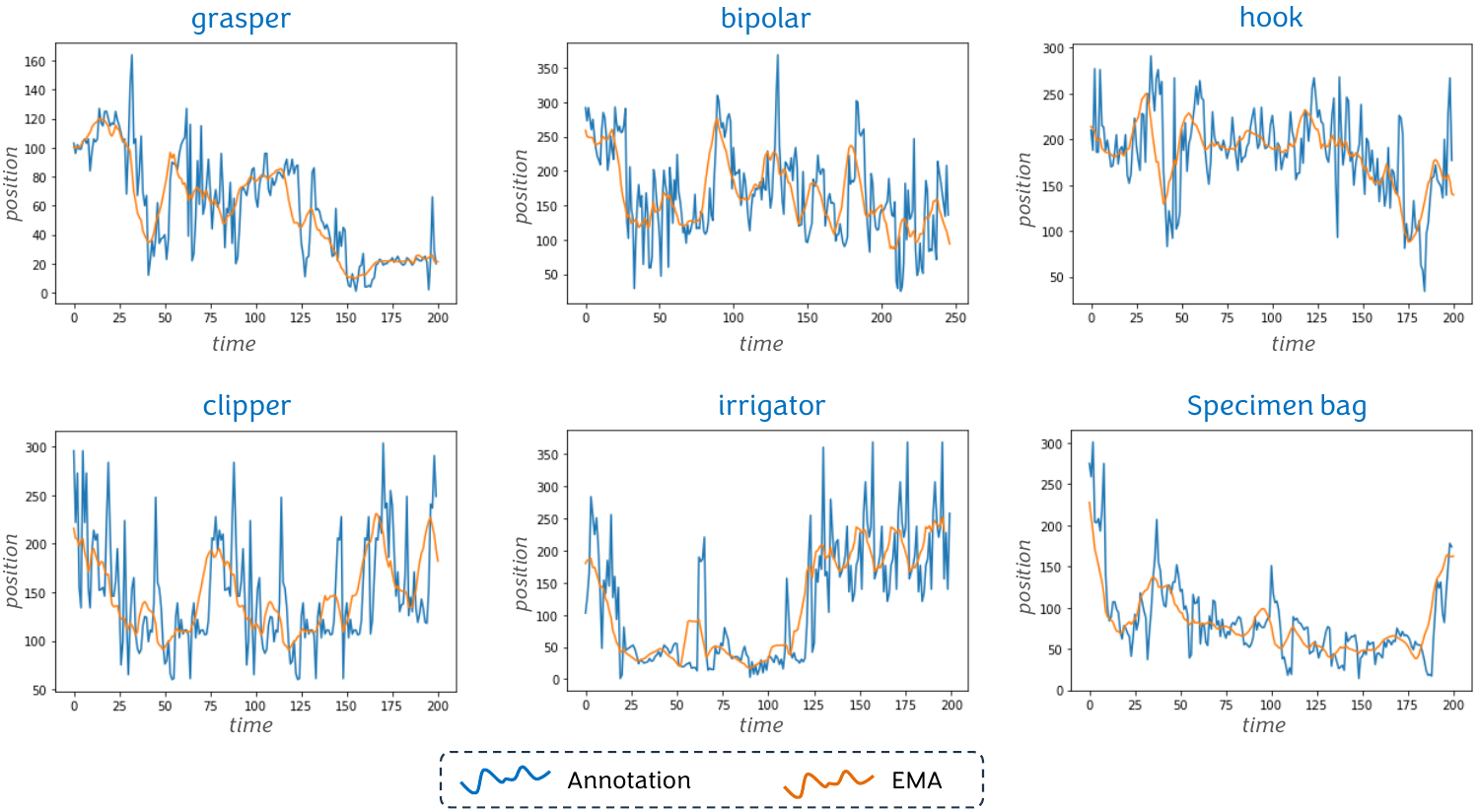}
    \caption{Trajectories of selected tools with EMA over time. Plotted positions are computed as weighted combinations of the center coordinate of the bounding boxes, scaled by image size.}
    \label{fig:label-ema}
\end{figure}

We conduct a comprehensive analysis of the dataset to gain insights into label alignments and feature similarities, revealing correlations across labels. This would guide data preprocessing and feature selection when using the data. 
Our analysis as illustrated in \cref{fig:label-alignment}, \cref{fig:phase-tracking}, and \cref{fig:label-ema} encompasses four distinct dimensions and discussed further.\smallskip

\noindent{\bf Tracking vs. surgical tool type correlation.} 
This analysis explores the relationship between different tool types and their tracks, providing insights into unique patterns associated with each tool during surgery. \cref{fig:label-alignment}(a) illustrates center point locations of tools, color-coded by category over time in three videos. This shows that while some trajectories appear separable, they are mostly densely interwoven, suggesting the need for advanced modeling in tool tracking.\smallskip

\begin{table*}[!t]
    \centering
    \caption{Benchmark Results of SOTA Object Detectors on Surgical Tool Detection Dataset.}
    \label{tab:benchmark-dets}
    \setlength{\tabcolsep}{2pt}
    \resizebox{\linewidth}{!}{%
        \begin{tabular}{@{}l rccc rccccccc rcccccccc rr@{}}
            \toprule
            \multirow{2}{*}{Detector} & \phantom{abc} &\multicolumn{3}{c}{Detection AP accross 3 thresholds} &\phantom{abc} & \multicolumn{7}{c}{Detection AP per category. (\% AP @ $\Theta=0.5$)} &\phantom{abc} & \multicolumn{8}{c}{Detection AP across surgical visual challenges} &\phantom{abc} & {Speed} \\
            \cmidrule{3-5}  \cmidrule{7-13}  \cmidrule{15-22}            
            Model && $AP_{0.5}\uparrow$ & $AP_{0.75}\uparrow$ & $AP_{0.5:0.95}\uparrow$ && Grasper & Bipolar & Hook & Scissors & Clipper & Irrigator & Bag && Bleeding & Blur & Smoke & Crowded & Occluded & Reflection & Foul Lens & Trocar  && FPS$\uparrow$  \\ 
            \midrule

            Faster-RCNN \cite{Ren2017faster} && 56.0 & 38.1 & 34.6 && 53.5 & 65.0 & 80.1 & 60.9 & 70.1 & 26.8 & 31.8 && 57.9 & 41.0 & 54.5 & 43.5 & 55.0 & 46.9 & 41.2 & 35.7 && 7.6 \\
            Cascade-RCNN \cite{Cai2019cascade} &&  51.7 & 39.0 & 34.7 && 52.0 & 58.9 & 79.7 & 45.7 & 44.9 & 23.7 & 17.9 && 53.9 & 39.0 & 48.1 & 39.5 & 46.4 & 29.1 & 33.7 & 33.7 && 7.0 \\
            CenterNet \cite{zhou2019centernet} && 53.0 & 39.5 & 35.0 && 60.2 & 61.4 & 86.4 & 56.3 & 68.0 & 25.8 & 10.2 && 58.0 & 42.1 & 50.2 & 36.7 & 51.7 & 46.0 & 35.8 & 30.8 && \bf 33.8 \\
            FCOS \cite{tian2019fcos} &&  43.5 & 31.5 & 28.1 && 51.2 & 44.3 & 74.7 & 49.2 & 54.2 & 21.9 & 7.2 && 47.8 & 40.6 & 51.5 & 15.1 & 40.8 & 42.7 & 29.7 & 17.6 && 7.7  \\
            SSD \cite{Liu2016ssd} &&  61.9 & 37.8 & 36.1 && 75.2 & 62.2 & 91.6 & 63.4 & 72.9 & 22.5 & 40.8 && 64.5 & 49.3 & 58.3 & 57.5 & 62.4 & 53.9 & 47.7 & 42.6 && 30.9 \\
            PAA \cite{paa-eccv2020} &&  64.5 & 44.9 & 41.1 && 69.6 & 79.0 & 89.2 & 68.7 & 74.2 & 37.6 & 28.9 && 67.1 & 55.6 & 65.0 & 55.0 & 64.6 & 56.0 & 51.2 & 47.5 && 8.5 \\
            Def-DETR \cite{zhu2021deformable} &&  58.4 & 42.0 & 38.3 && 60.6 & 66.5 & 83.8 & 61.9 & 72.0 & 39.9 & 23.8 && 62.4 & 42.7 & 58.6 & 37.1 & 57.4 & 43.9 & 41.5 & 47.4 && 10.2 \\
            Swin-T  \cite{liu2021Swin} &&  62.3 & 44.3 & 40.2 && 63.3 & 64.8 & 83.0 & 80.2 & 77.2 & 38.0 & 26.8 && 63.5 & 53.8 & 62.8 & 35.3 & 61.1 & 66.2 & 55.2 & 45.7 && 9.8 \\
            YOLOX \cite{yolox2021} &&  64.7 & 48.9 & 44.2 && 69.6 & 72.2 & 89.4 & 75.4 & 79.1 & 37.3 & 27.1 && 68.2 & 55.6 & 66.0 & 45.9 & 64.2 & 52.5 & 58.1 & 43.1 && 23.6 \\
            YOLOv7 \cite{wang2023yolov7} && \bf 80.6 & 62.0 & 56.1 && \bf 90.5 & 86.4 & 96.0 & \bf 82.3 & \bf 89.3 & 49.1 & 66.2  && \bf 80.2 & 61.2 & \bf 80.1 & \bf 79.5 & \bf 82.1 & 65.6 & \bf 71.2 & \bf 66.7 && 20.6 \\
            YOLOv8 \cite{varghese2024yolov8} && 79.1 & 62.4 & 55.6 && 87.9 & 84.5 & \bf 96.2 & 80.0 & 87.2 & 48.4 & 65.0 &&  77.1 & 58.3 & 74.4 & 76.2 & 80.4 & \bf 70.3 & 57.4 & 62.9 && 29.0 \\
            YOLOv9 \cite{wang2024yolov9} && 80.2 & \bf 62.6 & \bf 56.5 && 88.5 & \bf 87.6 & 96.0 & 79.3 & 87.1 & 50.1 & \bf 67.7 && 78.1 & 54.0 & 78.2 & 78.6 & 81.1 & 65.3 & 63.4 & 63.1 && 23.7 \\
            YOLOv10 \cite{wang2024yolov10} && 80.1 & 62.1 & 55.8 && 87.6 & 86.6 & 96.0 & 81.9 & 89.0 & \bf 53.8 & 61.3 && 77.8 & \bf 61.9 & 78.7 & 77.5 & 81.2 & 66.7 & 59.3 & 65.4 && 28.6 \\
            \bottomrule
            
        \end{tabular}
    }%
\end{table*}

\noindent{\bf Tracking vs. surgeon tool operator correlation.} We analyzed the alignment between tool operator identities and tool locations, revealing dynamic interactions between surgeons and instruments. \cref{fig:label-alignment}(b) shows a strong correlation between operators and their tools, attributed to the distinct positioning of trocars. This underscores the value of operator information in accurate track label generation.
\smallskip


\noindent{\bf Tracking vs. surgical phase segmentation.} 
This analysis delves into tracking tools across surgical phases, uncovering how tool utilization varies with procedural stages. \cref{fig:phase-tracking} shows that complex phases, such as calot triangle and gallbladder dissection, exhibit densely packed trajectories due to prolonged duration and frequent tool manipulation. Simpler phases, like preparation and clipping, feature fewer trajectories, facilitating modeling. These insights inform deep learning model assessment and AI model development for surgical tool tracking.\smallskip

\noindent{\bf Tracking variance.} 
Using Exponential Moving Average (EMA), we analyze tool tracking data, as shown in \cref{fig:label-ema}. By overlaying EMAs on tool trajectories, we visualize variance between actual trajectories and modeled tracks. This approach highlights high-variance frames, serving as challenging cases for benchmarking model robustness and accuracy.
Through this analysis, we provide a valuable resource for researchers, emphasizing the need to focus on complex scenarios for a rigorous assessment of model capabilities.

\section{Benchmark and Experiments}
\label{sec:experiments}

\subsection{Tool Detection}
\noindent{\bf Models.}
Owing that tool detection is a fundamental part of tool tracking, we showcase the usability of the CholecTrack20 dataset for this task, by benchmarking several object detectors representing diverse methodologies. Faster-RCNN \citep{Ren2017faster} and Cascade-RCNN \citep{Cai2019cascade} are anchor-based models, with Cascade-RCNN employing a multi-stage approach to refine detection accuracy. CenterNet \citep{zhou2019centernet} and FCOS \citep{tian2019fcos} are anchor-free models, utilizing center points and direct bounding box regression for efficient detection. SSD \citep{Liu2016ssd} provides real-time performance with its multi-scale approach. Deformable-DETR \citep{zhu2021deformable} applies a transformer-based method for flexible feature processing, while Swin-T \citep{liu2021Swin} uses hierarchical transformers with shifted windows. The YOLO models \cite{yolox2021,wang2023yolov7,varghese2024yolov8,wang2024yolov9,wang2024yolov10} feature advanced multi-scale strategies for high accuracy and speed.\smallskip


\noindent{\bf Evaluation metrics.}
We evaluate tool detection using COCO standard average precision (AP) metrics (pycocotools, {\it not ultralytics}) across several thresholds, categories, and visual challenges. We also report model inference speed in frames per seconds (FPS) on a single NVIDIA GTX 1080 Ti (10 GB) GPU.
\smallskip

\noindent{\bf Overall tool detection results.}
The detection performance of the models is summarized in \cref{tab:benchmark-dets}. YOLOv7 \citep{wang2023yolov7} achieves the highest Average Precision (AP) of 56.1\%, surpassing other models. YOLOX \citep{yolox2021} follows with an AP of 44.2\%. Deformable-DETR \citep{zhu2021deformable} and Swin-T \citep{liu2021Swin} show competitive results with APs of 38.3\% and 40.2\%, respectively.
The anchor-free models such as CenterNet \citep{zhou2019centernet} and FCOS \citep{tian2019fcos} demonstrate robust performance, with CenterNet achieving an AP of 35.0\% and FCOS 28.1\%. In comparison, Faster-RCNN \citep{Ren2017faster} and Cascade-RCNN \citep{Cai2019cascade} deliver APs of 34.6\% and 34.7\%, respectively, showcasing their efficacy with anchor-based approaches.
At IoU thresholds of 0.5 and 0.75, YOLOv7 leads with scores of 80.6\% and 62.0\%, respectively. In terms of inference speed, YOLO networks excel with real-time capacities exceeding 20 FPS, with CenterNet achieving the highest speed at 33.8 FPS.\smallskip

\noindent{\bf Class-wise detection results.}
Analyzing tool detection results per category (\cref{tab:benchmark-dets}), YOLOv7 emerges as the top performer, dominating in all the 7 categories, achieving above 90\% accuracy in 2 tool categories and above 80\% in 5. 
Notably, the hook exhibits the highest tendency among tools, with AP scores ranging from 74.7\% to 96.0\% across all models. Conversely, irrigator and specimen bag pose challenges, likely due to unclear tool tip boundaries and the bag's deformable nature, respectively. Grasper, bipolar, scissors, and clipper show high detection rates.\smallskip

\begin{table*}[!t]
    \centering
    \caption{Benchmark Multi-Perspective Multi-Tool Tracking Results @ 25 FPS.}
    \label{tab:benchmark-trackers}
    \setlength{\tabcolsep}{3pt}
    \resizebox{0.95\linewidth}{!}{%
    \begin{tabular}{@{}lcccclccccclccclcclr@{}}
        \toprule
        Model & 
        \multicolumn{4}{c}{HOTA Metrics} &\phantom{abc}&
        \multicolumn{5}{c}{CLEAR Metrics} &\phantom{abc}&
        \multicolumn{3}{c}{Identity Metrics} &\phantom{abc}&
        \multicolumn{2}{c}{Count Metrics} &\phantom{abc}&
        \multicolumn{1}{c}{Speed} \\        
        \cmidrule{2-5}
        \cmidrule{7-11}
        \cmidrule{13-15}
        \cmidrule{17-18}
        \cmidrule{20-20}
        Tracker 
        & HOTA$\uparrow$ 
        & DetA$\uparrow$
        & LocA$\uparrow$ 
        & AssA$\uparrow$
        &        
        & MOTA$\uparrow$
        & MOTP$\uparrow$
        & MT$\uparrow$
        & PT$\downarrow$
        & ML$\downarrow$
        &
        & IDF1$\uparrow$ 
        & IDSW$\downarrow$
        & Frag$\downarrow$
        &
        & \#Dets
        & \#IDs
        &
        & FPS$\uparrow$ \\
        \midrule \midrule
        
        &\multicolumn{18}{c}{\normalsize\bf\it Intraoperative Trajectory (Groundtruth counts: \#Dets = 29994, \#IDs = 70)}\\
        \midrule
        OCSORT \cite{maggiolino2023deep} & 14.6 & 52.7 & \bf 86.7 & 4.1 && 49.2 & \bf 85.0 & 24 & 32 & 14 && 9.5 & 2921 & 2731 && 21936 & 3336 && 10.2\\        
        FairMOT \cite{zhang2021fairmot}& 5.8 & 25.8 & 75.9 & 1.3 && 5.0 & 73.9 & 3 & 24 & 43 && 4.3 & 4227 & 1924 && 15252 & 4456 && 14.2 \\        
        TransTrack \cite{sun2020transtrack} & 7.4 & 31.5 & 84.4 & 1.7 && 4.2 & 82.9 & 9 & 36 & 25 && 4.2 & 4757 & \bf 1899 && 21640 & 4079 && 6.7\\        
        ByteTrack \cite{zhang2022bytetrack} & 15.8 & 70.6 & 85.7 & 3.6 && 67.0 & 84.0 & 54 & 12 & 2 && 9.5 & 4648 & 2429 && 28440 & 5383 && 16.4 \\        
        Bot-SORT \cite{aharon2022bot} & 17.4  & 70.7 & 85.4 & 4.4 && 69.6 & 83.7 & \bf 58 & 11 & \bf 1 && 10.2 & 3907 & 2376 && 29302 & 4501 && 8.7 \\        
        SMILETrack \cite{wang2023smiletrack} & 15.9 & 71.0 & 85.5 & 3.7 && 66.4 & 83.8 & 55 & 13 & 2 && 9.2 & 4968 & 2369 && 28821 & 5761 && 11.2 \\
        \midrule \midrule
        
        &\multicolumn{18}{c}{\normalsize\bf\it Intracorporeal Trajectory (Groundtruth counts: \#Dets = 29994, \#IDs = 247)}\\
        \midrule
        OCSORT \cite{maggiolino2023deep} & 23.7 & 51.4 & 86.5 & 11.0 && 47.1 & 84.8 & 115 & 87 & 45 && 18.1 & 2953 & 2796 && 21797 & 3526 && 10.2 \\
        FairMOT \cite{zhang2021fairmot}& 7.5  & 19.7 & 76.1 & 2.9 && 5.4 & 74.0 & 19 & 60 & 168 && 6.0 & 2890 & 1496 && 11287 & 3962 && 14.2 \\
        TransTrack \cite{sun2020transtrack} & 13.1  & 31.5 & 84.4 & 5.5 && 4.6 & 82.9 & 80 & 79 & 88 && 8.7 & 4648 & \bf 1791 && 21640 & 4079 && 6.7 \\
        ByteTrack \cite{zhang2022bytetrack} & 24.7  & 70.6 & 85.7 & 8.7 && 67.4 & 84.0 & 176 & 48 & 23 && 16.9 & 4515 & 2290 && 28440 & 5383 && 16.4 \\
        Bot-SORT \cite{aharon2022bot} & 27.0 & 70.7 & 85.4 & 10.4 && 70.0 & 83.7 & \bf 188 & 38 & \bf 21 && 18.9 & 3771 & 2238 && \bf 29300 & 4501 && 8.7 \\
        SMILETrack \cite{wang2023smiletrack} & 24.9  & 66.7 & 85.5 & 8.9 && 66.7 & 83.8 & 186 & 39 & 22 && 16.9 & 4868 & 2232 && 28820 & 5779 && 11.2 \\
        \midrule \midrule
        
        &\multicolumn{18}{c}{\normalsize\bf\it Visibility Trajectory (Groundtruth counts: \#Dets = 29994, \#IDs = 916)}\\
        \midrule
        SORT \cite{bewley2016simple}& 17.4 & 39.5 & 85.2 & 7.8 && 21.4 & 83.3 & 139 & 399 & 378 && 13.4 & 6619 & 2138 && 16595 & 8844 && 19.5 \\        
        OCSORT \cite{maggiolino2023deep} & 37.0 & 52.6 & 86.5 & 26.2 && 50.2 & 84.8 & 300 & 371 & 245 && 35.9 & 2317 & 2260 && 22197 & 3587 && 10.2 \\
        FairMOT \cite{zhang2021fairmot}& 15.3 & 25.0 & 75.8 & 9.5 && 7.1 & 73.7 & 58 & 218 & 640 && 14.4 & 3140 & 1574 && 15338 & 4875 && 14.2 \\
        TransTrack \cite{sun2020transtrack} & 19.2 & 31.6 & 84.4 & 11.8 && 5.8 & 82.9 & 224 & 280 & 412 && 16.1 & 4273 & \bf 1403 && 21640 & 4079 && 6.7 \\
        ByteTrack \cite{zhang2022bytetrack} & 41.5 & 70.7 & 85.7 & 24.8 && 69.3 & 84.0 & 591 & 217 & 108 && 36.8 & 3930 & 1704 && 28440 & 5383 && 16.4 \\
        Bot-SORT \cite{aharon2022bot} & 44.7 & 70.8 & 85.5 & 28.7 && 72.0 & 83.7 & \bf 638 & 184 & \bf 94 && 41.4 & 3183 & 1638 && \bf 29300 & 4505 && 8.7 \\
        SMILETrack \cite{wang2023smiletrack} & 41.3 & 71.0 & 85.6 & 24.4 && 68.9 & 83.8 & 619 & 192 & 105 && 36.5 & 4227 & 1641 && 28821 & 5752 && 11.2 \\
        \bottomrule
    \end{tabular}
    }
\end{table*}

\noindent{\bf Detection under visual challenges.}
\cref{tab:benchmark-dets} presents the performance of the benchmark object detection models across different surgical visual challenges. Notably, YOLOv7 achieves the highest detection accuracy across most challenges, particularly excelling in scenarios involving bleeding, smoke, and crowded scenes. Conversely, detecting tools in blurred scenes, near trocars, and specular light reflection pose significant challenges for all models, with lower detection rates observed across the board.

\subsection{Tool Tracking}
\noindent{\bf Models.}
We train and evaluate several state-of-the-art multi-object tracking (MOT) methods on the CholecTrack20 dataset, focusing on their ability to track surgical tools. OCSORT \citep{maggiolino2023deep} and TransTrack \citep{sun2020transtrack} employ sophisticated tracking-by-detection frameworks, with TransTrack utilizing transformers to improve feature association. ByteTrack \citep{zhang2022bytetrack}, Bot-SORT \citep{aharon2022bot}, and SMILETrack \citep{wang2023smiletrack} use advanced tracking algorithms, with Bot-SORT and SMILETrack incorporating extra modules for enhanced robustness. 

\noindent{\bf Evaluation metrics.}
We assess benchmark models on variety of tracking metrics: higher-order tracking accuracy (HOTA) \citep{luiten2021hota}, CLEAR MOT metrics \citep{bernardin2008evaluating}, identity metrics \citep{ristani2016performance}, counting metrics, and tracking speed.
A pull request is made to the standard TrackEval \cite{luiten2020trackeval} library integrating CholecTrack20 benchmark with all its exhaustive performance evaluation protocols recommended by this study.
\smallskip

\noindent{\bf Multi-object tracking results.}
The performance of the evaluated tracking methods is summarized in \cref{tab:benchmark-trackers}. Models such as FairMOT and TransTrack show the lowest HOTA score of 5.8\% and 7.4\% respectively, highlighting challenges in maintaining tool identities over time. ByteTrack, Bot-SORT, and SMILETrack achieve higher HOTA scores, ranging from 15.7\% to 17.4\%, but still face difficulties in tool re-ID due to similarities among tools. 

Despite these advancements, there remains room for enhancement in identity association and tracking precision. The results also include metrics on the detection counts, unique identities assigned, and tracking speed, providing a comprehensive view of each method’s performance.
\smallskip

\noindent{\bf Multi-perspective tracking results.}
Looking at the different trajectory perspectives, \cref{tab:benchmark-trackers} shows that visibility tracking is the easiest with most of the existing models showcasing their strengths. This is expected because deep learning models mostly rely on visual cues, which are captured by camera in the visibility track scenario. 
Here, Bot-SORT record a landslide top performance scores of 44.7\% HOTA, 72.0\% MOTA, and 41.4\% IDF1.
The intracorporeal tracking is the most challenging since the major factors marking the entry and exit of the tools from the body are not readily visible. A maximum of 27.0\% HOTA suggest a decline on the leading Bot-SORT. New methods could leverage rich fine-grained history to estimate the out-of-view and out-of-body status of the tools for improve re-ID.
The intraoperative trajectory comes in the middle in terms of difficulty. While it may be challenging to ascertain the persistence of a trajectory after re-insertion, the class features are also helpful especially for tools of different categories. Again the Bot-SORT, leverage camera compensation details, shows a better tendency of estimating the persistent identity of different tools of the same class with a +1.5\% HOTA higher than similarity and appearance features. 
\smallskip

\begin{figure}[!t]
    \centering
    \includegraphics[width=.99\columnwidth]{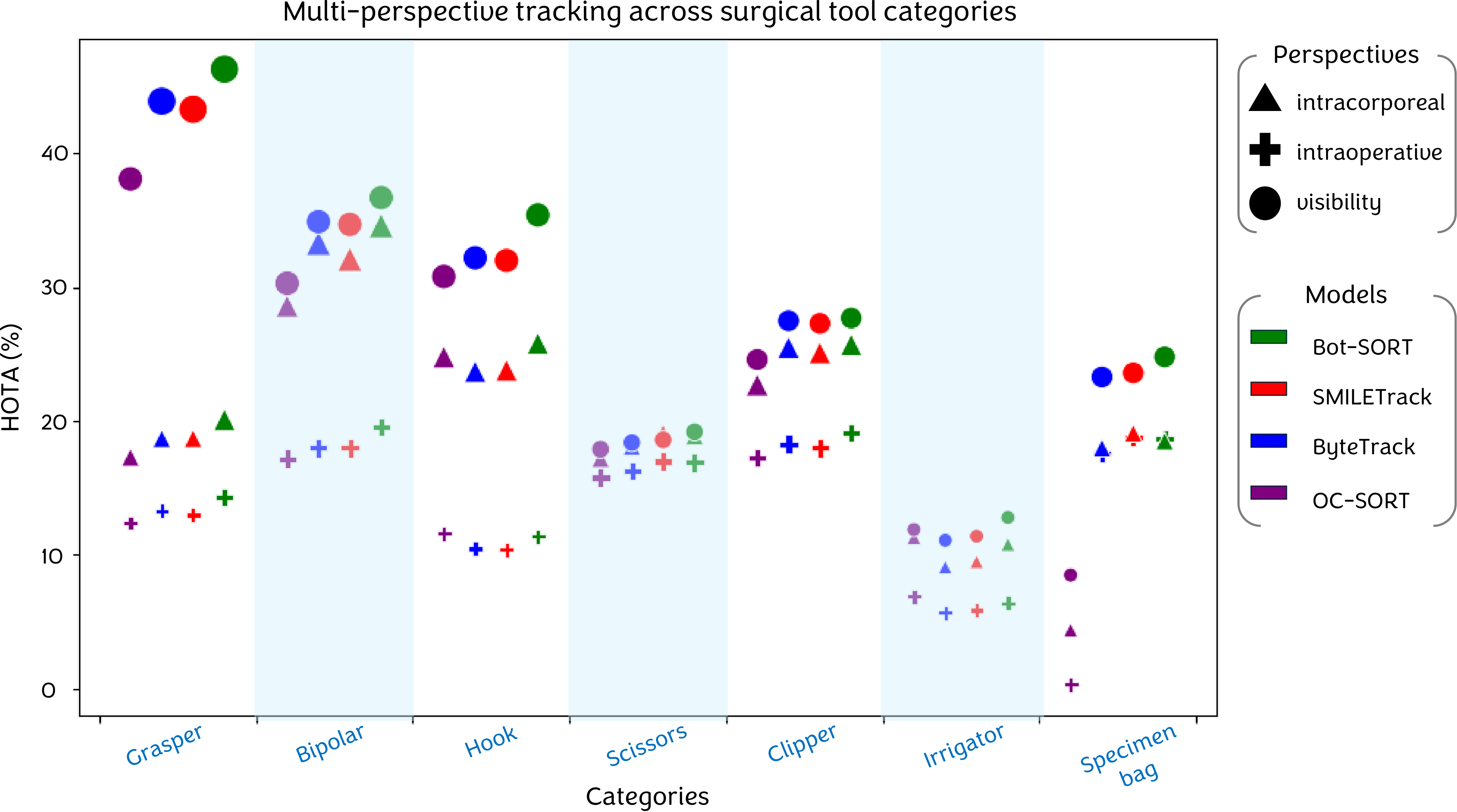}
    \caption{Results of multi-perspective tracking across seven surgical tool categories in the CholecTrack20 dataset.}
    \label{fig:results-mc}
\end{figure}

\noindent{\bf Multi-class tracking results.}
In \cref{fig:results-mc}, we analyze tracking performance by tool class and observe that the grasper, despite having the most instances and being the most frequently used tool in the dataset, achieves the highest tracking accuracy across perspectives. Class-agnostic results reveal medium tracking accuracy for other commonly used tools like the bipolar, hook, and clipper, while rarely used tools (e.g., scissors, irrigator) have lower scores. 
%
Specimen bag tracking is affected by shape deformation, contents, states
(open/closed, empty/filled), tool interactions, and fluid stains.
Tracking surgical tools beyond the visual perspective remains challenging for all models tested.
\smallskip

\begin{figure}[!t]
    \centering
    \includegraphics[width=.99\columnwidth]{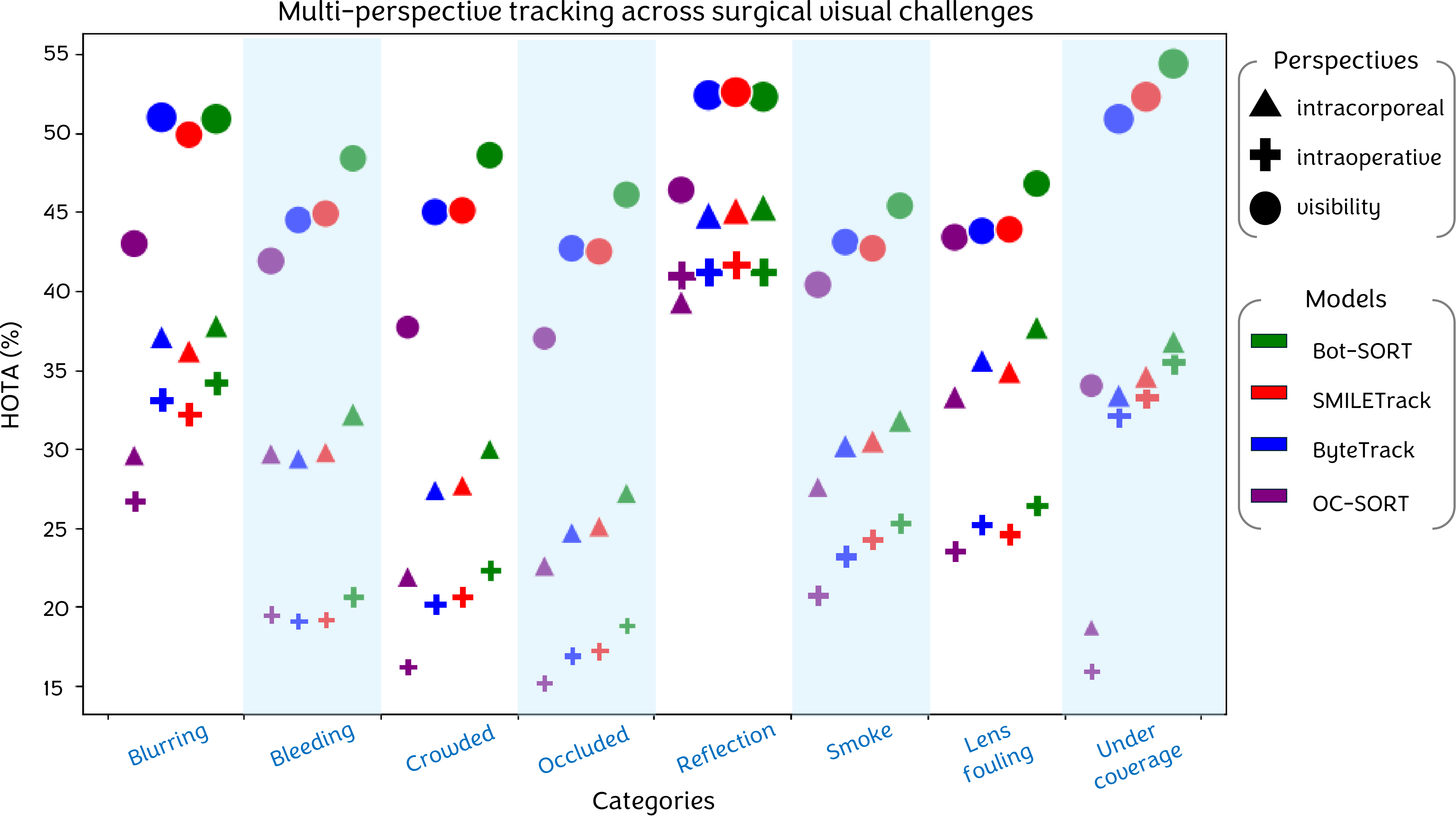}
    \caption{Results of multi-perspective tracking across eight surgical visual challenges in the CholecTrack20 dataset.}
    \label{fig:results-vc}
\end{figure}

\begin{figure}[ht]
    \centering
    \includegraphics[width=.99\columnwidth]{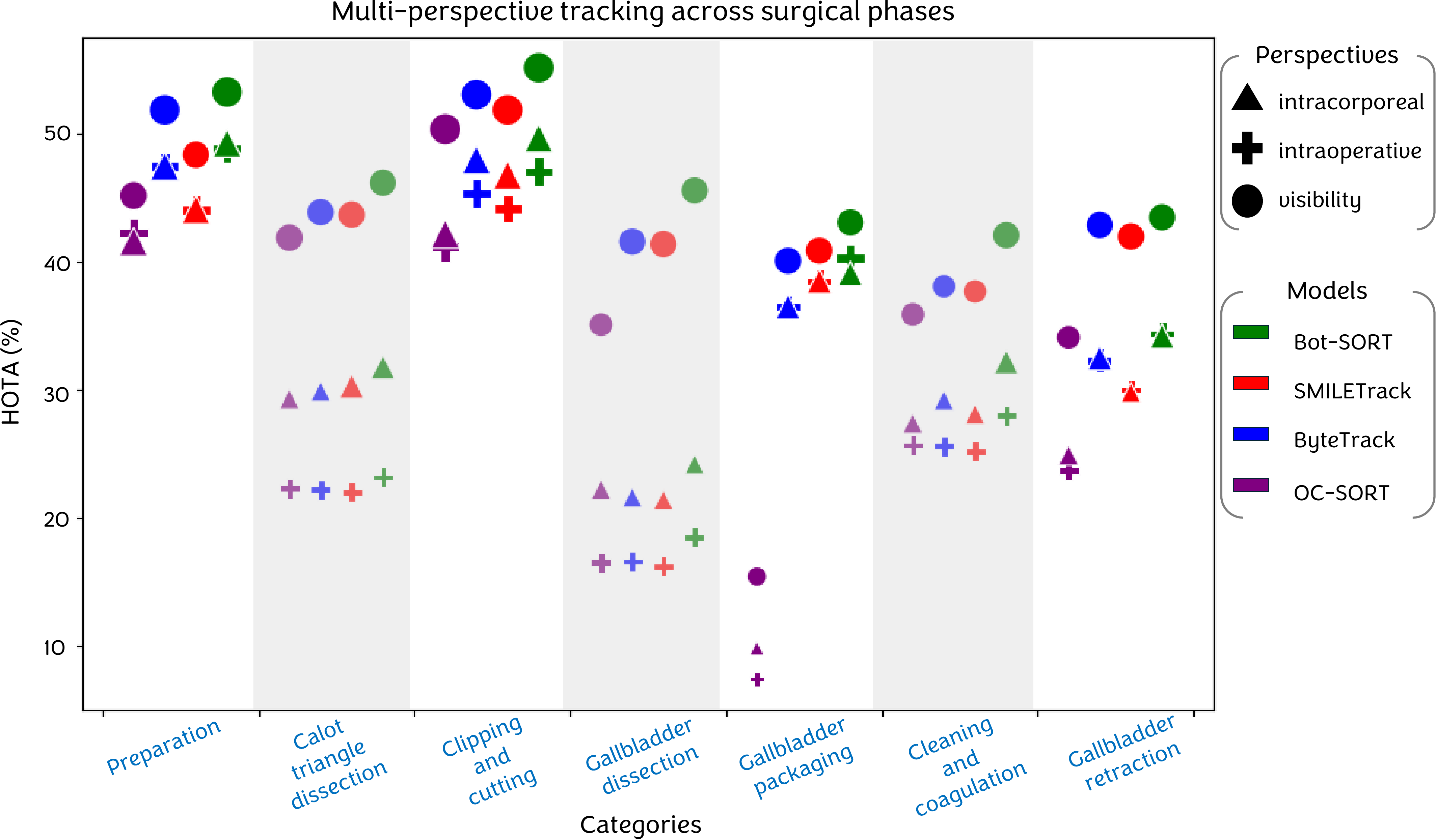}
    \caption{Results of multi-perspective tracking across seven surgical phases in the CholecTrack20 dataset.}
    \label{fig:results-phase}
\end{figure}

\noindent{\bf Tracking results under visual challenges.}  
We evaluate tool tracking performance across various surgical conditions using HOTA metrics in \cref{fig:results-vc}, providing insights into model interactions with complex surgical environments. The model performs well under blurring, reflections, and limited camera coverage, likely due to effective data augmentation. However, lens fouling, smoke, and occlusion present significant challenges, reducing accuracy.
\smallskip

\noindent{\bf Tracking results by surgical phase.}  
\cref{fig:results-phase} shows performance across seven surgical phases, with the clipping and cutting phases proving easiest to track due to limited activities and a linear progression. Preparation phase shows similar performance. Phases like Calot triangle dissection and gallbladder dissection exhibit comparable tracking results, while gallbladder packaging shows the most consistent tracking across perspectives. Overall, OC-SORT struggles the most, while Bot-SORT achieves the best result. 

\subsection{Limitations and Gaps to Address}
The SOTA tracking methods trained on CholecTrack20 reveal substantial limitations, with performance under 45\% HOTA, which is insufficient for clinical translation. These models struggle with various visual challenges, such as smoke, bleeding, and specular light reflection, affecting detection and re-ID. Since location and appearance features alone are inadequate, especially for tools with similar appearances, this highlights the need to move beyond current cues and innovate more intuitive, context-aware methods for re-ID. CholecTrack20 serves as a critical foundation for exploring this direction, offering a dataset rich in diverse tracking perspectives and challenges, essential for developing more robust and clinically viable tracking solutions.
\section{Conclusion}
\label{sec:conclusion}

In this work, we presented the CholecTrack20 dataset, a novel resource designed to advance the state-of-the-art in surgical tool tracking within computer vision. CholecTrack20 addresses a critical gap by providing comprehensive annotations and diverse tracking scenarios and tasks across various surgical phases and visual challenges.
Key innovations of CholecTrack20 include multi-perspective tracking, which defines the start and termination of a tool track differently based on visibility, intracorporeal, or intraoperative contexts. It also features detailed annotations of surgical visual challenges and precise surgical phase segmentation. 
Our extensive benchmark experiments demonstrate the dataset's effectiveness in developing models for tool detection and multi-object tracking across these three distinct trajectory perspectives. Through evaluating several deep learning methodologies on the CholecTrack20 dataset, we gain insights into their strengths and weaknesses in handling multiple viewpoints or tracking perspectives, across surgical phases and diverse surgical scene visual challenges.

By introducing CholecTrack20 to the computer vision research community, we aim to stimulate new research directions and foster collaborations between the computer vision and surgical communities. This dataset serves as a benchmark for evaluating state-of-the-art algorithms and promotes the development of robust and reliable surgical assistance systems.
We anticipate that CholecTrack20 will inspire innovative approaches and contribute significantly to advancements in surgical tool tracking and related fields.
The dataset is released under the CC BY-NC-SA license.

\smallskip

\subsection*{Acknowledgments}
This work was supported by French state funds managed within the Plan Investissements d’Avenir by the ANR under references: National AI Chair AI4ORSafety [ANR-20-CHIA-0029-01], DeepSurg [ANR-16-CE33-0009], IHU Strasbourg [ANR-10-IAHU-02] and by BPI France under references: project CONDOR, project 5G-OR. 
Joël L. Lavanchy received funding by the Swiss National Science Foundation (P500PM\_206724, P5R5PM\_217663). 
This work was granted access to the servers/HPC resources managed by CAMMA, IHU Strasbourg, Unistra Mesocentre, and GENCI-IDRIS [Grant 2021-AD011011638R3, 2021-AD011011638R4].

{
    \small
    \bibliographystyle{ieeenat_fullname}
    \bibliography{main}
}


\end{document}